\pdfoutput=1

\documentclass[11pt]{article}

\usepackage{ACL2023}

\usepackage{times}
\usepackage{latexsym}
\usepackage{graphicx}
\usepackage{makecell}
\usepackage{tabularx}
\usepackage{multirow}
\usepackage{adjustbox}
\usepackage{mathtools}
\usepackage{enumitem}
\usepackage{algorithm2e}
\usepackage{csquotes}
\usepackage{courier}
\usepackage{pgf}
\usepackage{amsmath}
\usepackage{microtype}
\usepackage{dblfloatfix}
\usepackage{hyperref}

\usepackage[T1]{fontenc}

\usepackage[utf8]{inputenc}

\usepackage{microtype}
\usepackage{printlen}

\title{How Good is the Model in Model-in-the-loop Event Coreference Resolution Annotation?}

\author{Shafiuddin Rehan Ahmed$^{1}$\hspace{0.5em} 
Abhijnan Nath$^{2}$\hspace{0.5em} 
Michael Regan$^{3}$\hspace{0.5em} \\ 
\textbf{Adam Pollins}$^{1}$ \hspace{0.5em} 
\textbf{Nikhil Krishnaswamy}$^{2}$ \hspace{0.5em}
\textbf{James H. Martin}$^{1}$ 
\vspace{3pt}\\  
    $^1${University of Colorado, Boulder, CO, USA} \hspace{3pt}
    $^{3}${University of Washington, Seattle, WA, USA } \\
    $^2${Colorado State University, Fort Collins, CO, USA} \\ 
    { \fontsize{10.00000}{10.800000}\selectfont \tt{\{shah7567,james.martin,adpo0729\}@colorado.edu}} \\
    {\fontsize{10.00000}{10.800000}\selectfont \tt {\{abhijnan.nath,nkrishna\}@colostate.edu}}, \hspace{3pt}
    {\fontsize{10.00000}{10.800000}\selectfont \tt {mregan@cs.washington.edu} } 
}

\begin{document}
\maketitle
\begin{abstract}
Annotating cross-document event coreference links is a time-consuming and cognitively demanding task that can compromise annotation quality and efficiency. To address this, we propose a model-in-the-loop annotation approach for event coreference resolution, where a machine learning model suggests likely corefering event pairs only. We evaluate the effectiveness of this approach by first simulating the annotation process and then, using a novel annotator-centric Recall-Annotation effort trade-off metric, we compare the results of various underlying models and datasets. We finally present a method for obtaining 97\% recall while substantially reducing the workload required by a fully manual annotation process.
\end{abstract}

\section{Introduction}
\newcommand{\mone}{\ensuremath{m_{1}}}
\newcommand{\mtwo}{\ensuremath{m_{2}}}
\newcommand{\mthree}{\ensuremath{m_{3}}}
\newcommand{\mfour}{\ensuremath{m_{4}}}

\newcommand{\eone}{\ensuremath{e_{1}}}
\newcommand{\etwo}{\ensuremath{e_{2}}}
\newcommand{\ethree}{\ensuremath{e_{3}}}

\newcommand{\ecbone}{\textbf{39\_11ecbplus}}
\newcommand{\ecbtwo}{\textbf{39\_1ecb}}
\newcommand{\ecbthree}{\textbf{39\_5ecbplus}}
\newcommand{\argzero}{$_{0}$}
\newcommand{\argone}{$_{1}$}
\newcommand{\replace}{\textcolor{blue}{$replace_{\mone}$}}
\newcommand{\replacing}{\textcolor{magenta}{$replacing_{\mtwo}$}}

\newcommand{\takesover}{\textcolor{blue}{$takes~over_{\mthree}$}}

\newcommand{\steppedinto}{\textcolor{blue}{$stepped~into_{\mfour}$}}

\newcommand{\blue}[1]{\textcolor{blue}{\textit{#1}}}
\newcommand{\evtone}{\blue{$e_{1}$}}
\newcommand{\evttwo}{\textcolor{magenta}{$e_{2}$}}
\newcommand{\evtthree}{\blue{$e_{3}$}}
\newcommand{\evtfour}{\blue{$e_{4}$}}

Event Coreference Resolution (ECR) is the task of identifying mentions of the same event either within or across documents.
Consider the following excerpts from three related documents:
\begin{itemize}[label={},leftmargin=*, itemsep=0em]
\item \eone: 55 year old star will \replace~Matt Smith, who announced in June that he was leaving the sci-fi show.
  \item \etwo: Matt Smith, 26, will make his debut in 2010, \replacing~David Tennant, who leaves at the end of this year.
  \item \ethree: Peter Capaldi \takesover~Doctor Wh\-o~\dots~Peter Capaldi~\steppedinto~Matt Smith's soon to be vacant Doctor Who shoes.
\end{itemize}

\noindent \eone, \etwo, and \ethree~are example sentences from three documents where the event mentions are highlighted and sub-scripted by their respective identifiers (\mone~through \mfour). The task of ECR is to automatically form the two clusters $\{\mone, \mthree, \mfour\}$, and $\{\mtwo\}$. We refer to any pair between the mentions of a cluster, e.g., $(\mone, \mthree)$ as an ECR link. Any pair formed across two clusters, e.g., $(\mone, \mtwo)$ is referred to as non-ECR link.






Annotating ECR links can be challenging due to the large volume of mention pairs that must be compared. The annotating task becomes increasingly time-consuming as the number of events in the corpus increases. As a result, this task requires a lot of mental effort from the annotator and can lead to poor quality annotations \cite{song-etal-2018-cross, wright-bettner-etal-2019-cross}. Indeed, an annotator has to examine multiple documents simultaneously often relying on memory to identify all the links which can be an error-prone process.

To reduce the cognitive burden of annotating ECR links, annotation tools can provide integrated model-in-the-loop for sampling likely coreferent mention pairs \cite{pianta-etal-2008-textpro,yimam-etal-2014-automatic,klie-etal-2018-inception}. These systems typically store a knowledge base (KB) of  annotated documents and then use this KB to suggest relevant candidates. The annotator can then inspect the candidates and choose a coreferent event if present.

The model's querying and ranking operations are typically driven by machine learning (ML) systems that are trained either actively \cite{pianta-etal-2008-textpro, klie-etal-2018-inception,bornstein-etal-2020-corefi,yuan-etal-2022-adapting} or by using batches of annotations \cite{yimam-etal-2014-automatic}. While there have been advances in suggestion-based annotations, there is little to no work in evaluating the effectiveness of these systems, particularly in the use case of ECR. Specifically, both the overall coverage, or recall, of the annotation process as well as the degree of annotator effort needed depend on the performance of the model. In order to address this shortcoming, we offer the following contributions:
\vspace*{-1.5mm}
\begin{enumerate}
\setlength{\itemsep}{0pt}
    \item We introduce a method of model-in-the-loop annotations for ECR\footnote{repo: \href{https://github.com/ahmeshaf/model_in_coref}{$\mathtt{github.com/ahmeshaf/model\_in\_coref}$}}.
    \vspace*{-0.5mm}
    \item We compare three existing methods for ECR (differing widely in their computational costs), by adapting them as the underlying ML models governing the annotations.
    \vspace*{-0.5mm}
    \item We introduce a novel methodology for assessing the workflow by simulating the annotations and then evaluating an annotator-centric Recall-Annotation effort tradeoff.
\end{enumerate}



\section{Related Work}



Previous work for ECR is largely based on modeling the probability of coreference between mention pairs. These models are built on supervised classifiers trained using features extracted from the pairs. 
Most recent work uses a transformer-based language model (LM) like BERT \cite{DBLP:journals/corr/abs-1810-04805, liu2019roberta} to generate joint representations of mention pairs, a method known as cross-encoding. The cross-encoder is fine-tuned using a coreference scoring objective~\citep{barhom-etal-2019-revisiting, cattan2020streamlining,meged-etal-2020-paraphrasing, zeng-etal-2020-event, yu2020paired, caciularu-etal-2021-cdlm-cross}. These methods use scores generated from the scorer to then agglomeratively cluster coreferent events. 

\newcommand{\bfunc}[1]{\mathone{\operatorname{\mathtt{#1}}}}

\newcommand{\mucr}{\bfunc{Recall}}
\newcommand{\preci}{\bfunc{P}}
\newcommand{\comps}{\bfunc{Comparisons}}
\newcommand{\n}{\mathone{k}}

Over the years, a number of metrics have been proposed to evaluate ECR~\citep{10.3115/1072399.1072405,Bagga98algorithmsfor, 10.3115/1220575.1220579, blanc, luo-etal-2014-extension, pradhan-EtAl:2014:P14-2}. An ECR system is evaluated using these metrics to determine how effectively it can find event clusters (recall) and how cleanly separated the clusters are (precision). From the perspective of annotation, it may only be necessary to focus on the system's recall or its effectiveness in finding ECR links. However, an annotator might still want to know how much effort is  required to identify these links in a corpus to estimate their budget. In the remainder of the paper, we attempt to answer this question by first quantifying annotation effort and analyzing its relation with recall of the system.

\newcommand{\ecb}{ECB+}
\newcommand{\gvc}{GVC}

We use the Event Coreference Bank Plus (\ecb; \citet{DBLP:conf/lrec/CybulskaV14}) and the Gun Violence Corpus (GVC; \citet{vossen2018don}) for our experiments. The ECB+
is a common choice for assessing ECR, as well as the  experimental setup of \citet{cybulska-vossen-2015-translating} and gold topic clustering of documents and gold mention annotations for both training and testing\footnote{The \ecb~test set has 1,780 event mentions with 5K ECR links among 100K pairwise mentions, while the \gvc~test set has 1,008 mentions with 2K ECR links in 20K pairs. Full statistics in Table \ref{tab:ecb_gvc} in Appendix \ref{sec:corpus_details}}. On the other hand, the GVC offers a more challenging set of exclusively event-specific coreference decisions that require resolving gun violence-related events. 

\newcommand{\events}{\textit{events}}
\newcommand\mycommfont[1]{\small\ttfamily\textcolor{blue}{#1}}
\SetCommentSty{mycommfont}
\newcommand{\eti}{\mathone{e^T_i}}
\newcommand{\thetabar}{\mathone{\overline{\thetaone}}}
\newcommand{\baseline}{\texttt{Baseline}}
\newcommand{\regressor}{\texttt{Regressor}}
\newcommand{\cdlm}{\texttt{CDLM}}
\newcommand{\cdlmT}{\ensuremath{\texttt{CDLM}_{\texttt{T}}}}
\newcommand{\bert}{\texttt{BERT}}
\newcommand{\lemma}{\texttt{Lemma}}
\newcommand{\manual}{\texttt{Random}}
\newcommand{\bertT}{\ensuremath{\texttt{BERT}_{\texttt{T}}}}
\newcommand{\edi}{\mathone{e^D_i}}
\newcommand{\mskipped}{\ensuremath{m_{*}}}
\begin{figure}[t] 
\begin{center}
\includegraphics[scale=0.55]{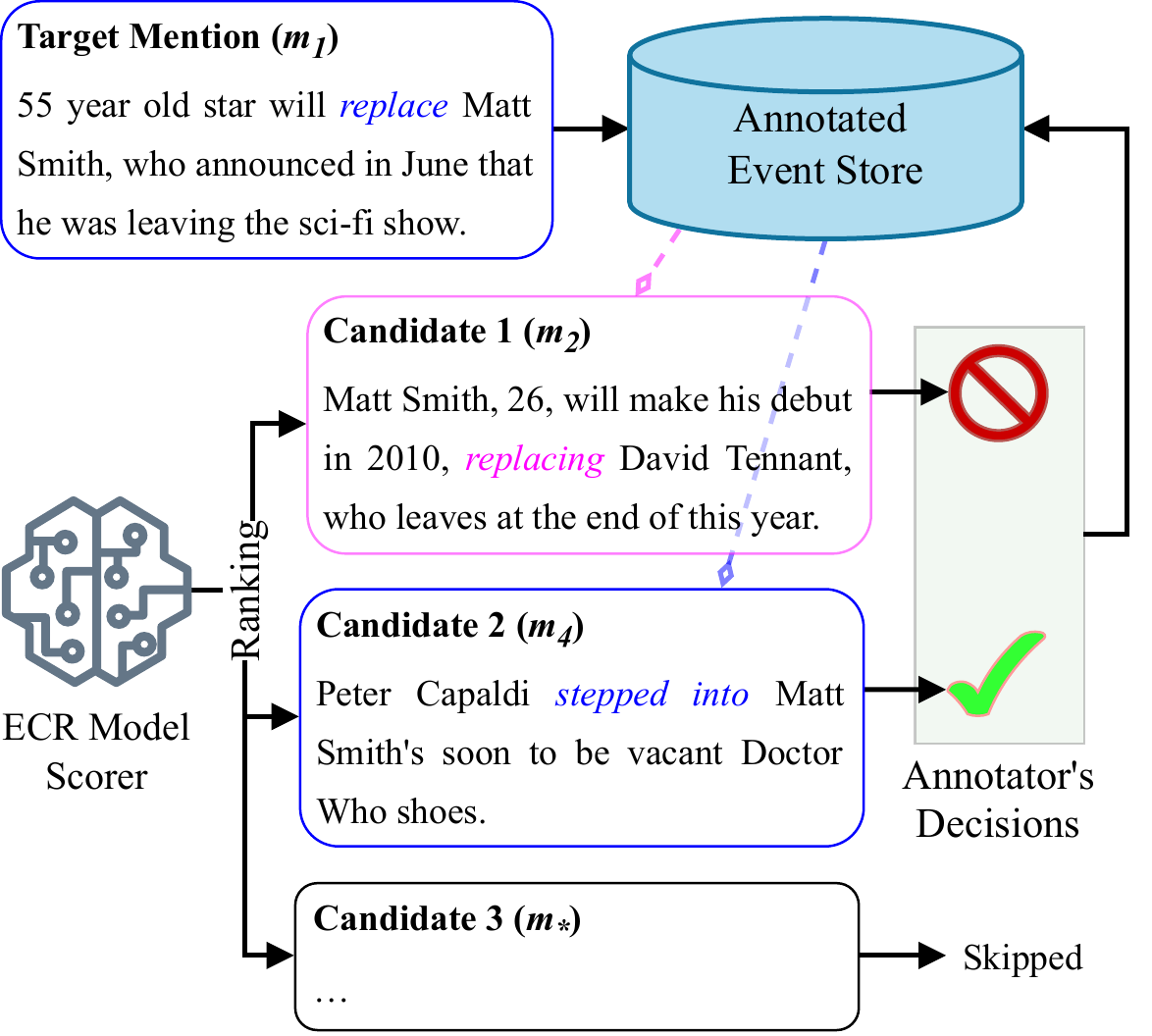}
\end{center}
\caption{For the target mention (\mone), the Annotated Event Cluster store presents three potential coreferent candidates (\mtwo, \mfour~and \mskipped). The ranking module (an ECR scorer) then ranks them based on their semantic similarity to \mone. The annotator reviews each candidate one-at-a-time and makes decisions on coreference. \mskipped~is skipped after finding \mfour~as coreferent. The cluster store is then updated based on these decisions.}
\label{fig:workflow}
\end{figure}
\vspace*{-10.5mm}
\section{Annotation Methodology}
\vspace*{-1.mm}
We implement an iterative model-in-the-loop meth\-odology\footnote{Utilizing the \href{https://prodi.gy}{prodi.gy} annotation tool. See Appendix \ref{app:prodigy}} for annotating ECR links in a corpus containing annotated event triggers. This approach has two main components - (1) the storage and retrieval of annotated event clusters, which are then compared with each new target event, and (2), an ML model that ranks and prunes the sampled candidate clusters by evaluating their semantic similarity to the target mention.

As illustrated in Figure \ref{fig:workflow}, our annotation workflow queries the Annotated Event Store for the target event (\mone), retrieving three potential coreferring candidates (\mtwo, \mskipped, and \mfour). The ranking module then evaluates these candidates based on their lexical and semantic similarities to \mone. The annotator then compares each candidate to the target and determines if they are coreferent. Upon finding a coreferent candidate, the target is merged into the coreferent cluster, and any remaining option(s) (\mskipped) are skipped.


\newcommand{\mathone}[1]{\ensuremath{#1}}
\newcommand{\ei}{\ensuremath{e_i}}
\newcommand{\thetaone}{\ensuremath{\Theta}}

\newcommand{\mymathrm}[1]{\mathone{\mathrm{#1}}}
\newcommand{\clusters}{\mathone{\mathcal{E}}}

\newcommand{\bertscore}[1]{\ensuremath{\mathtt{BS}(#1)}}
\newcommand{\JS}{\ensuremath{\mathtt{JS}}}
\newcommand{\BertScore}[1]{\ensuremath{\texttt{BERTScore}(#1)}}
\newcommand{\bertsent}[1]{\ensuremath{S_{\text{b}}(#1)}}
\newcommand{\msubi}{\ensuremath{m_{i}}}
\newcommand{\msubj}{\ensuremath{m_{j}}}
\newcommand{\tsubmi}{\ensuremath{t_{m_{i}}}}
\newcommand{\tsubmj}{\ensuremath{t_{m_{j}}}}
\newcommand{\Ssubmi}{\ensuremath{S_{m_{i}}}}
\newcommand{\Ssubmj}{\ensuremath{S_{m_{j}}}}
\newcommand{\Long}{\ensuremath{\mathtt{LF}}}
\newcommand{\ECLS}{\ensuremath{\text{E}_{\text{CLS}}}}
\newcommand{\Emsubi}{\ensuremath{\text{E}_{\msubi}}}
\newcommand{\Emsubj}{\ensuremath{\text{E}_{\msubj}}}
\vspace*{-0.5mm}
\subsection{Ranking}
\vspace*{-0.5mm}
We investigate three separate methods to drive the ranking of candidates distinguished by their computational cost. We use these methods to generate the average pair-wise coreference scores between mentions of the candidate and target events, then use these scores to rank candidates. We use a single RTX 3090 (24 GB) for running our experiments.

\vspace{1.5mm}
\noindent\textbf{Cross-encoder} (\cdlm): In this method, we use the fine-tuned cross-encoder ECR system of \citet{caciularu-etal-2021-cdlm-cross} to generate pairwise mention scores\footnote{This method is compute-intensive since the transformer's encoding process scales quadratically with the number of mentions.  Using the trained weights, running inference on the two test sets for our experiments takes approximately forty minutes to calculate the similarities of all the mention pairs. The weights are provided by \citet{caciularu-etal-2021-cdlm-cross} \href{https://drive.google.com/drive/folders/1bVSI3VPAvcIixrclkZyTHDXi7bamg93b?usp=sharing}{here}.}. Their state of the art system uses a modified Longformer \citep{beltagy2020longformer} as the underlying LM to generate document-level representations of the mention pairs (detailed in \S\ref{app:cdlm}). More specifically, we generate a unified representation (Eq. 1) of the mention pair (\msubi, \msubj) by concatenating the pooled output of the transformer (\ECLS), the outputs of the individual event triggers (\Emsubi, \Emsubj), and their element-wise product. Thereafter, pairwise-wise scores are generated for each mention-pair after passing the above representations through a Multi-Layer Perceptron ($\mathtt{mlp}$) (Eq. 2) that was trained using the gold-standard labels for supervision. 
\begin{align}
\label{eqn:ce}
&\Long(\msubi, \msubj) = \small \left<\ECLS,\Emsubi,\Emsubj, \Emsubi \odot \Emsubj\right>\\
\label{eqn:cdlm}
&\cdlm(\msubi,\msubj) = \mathtt{mlp}(\Long(\msubi, \msubj))
\end{align}
 \noindent\textbf{BERTScore} (\bert):
\cite{zhang2019bertscore} 
BERT-Score ($\mathtt{BS}$) is a NLP metric that measures pairwise text similarity by exploiting pretrained BERT models. It calculates cosine similarity of token embeddings with inverse document frequency weights to rate token importance and aggregates them into precision, recall, and F1 scores. This method emphasizes semantically significant tokens, resulting in a more accurate similarity score (details in \S\ref{app:bert}).
\newcommand{\Sbert}{\ensuremath{\mathtt{S}_{\text{bert}}}}
\begin{align}
\label{eqn:BertSent}
\Sbert(&m) =~\left<t_{m},~\text{[SEP]},~S_{m}\right>\\
\label{eqn:BERT}
\begin{split}
\bert(&\msubi,\msubj) = \lambda~\bertscore{\tsubmi, \tsubmj}\\
                     &+ (1 - \lambda)~\bertscore{\Sbert(\msubi), \Sbert(\msubj)}
\end{split}
\end{align}
To calculate the BERTScore between the mentions, we first construct a combined sentence ($\Sbert(m)$; \citet{shi2019simple}) for a mention ($m$) by concatenating the mention text ($t_m$) and its corresponding sentence ($S_m$), as depicted in Equation \ref{eqn:BertSent}. Subsequently, we compute the $\mathtt{BS}$~for each mention pair using $\Sbert(m)$~and $t_m$ separately, then extract the F1 from each. We then take the weighted average of the two scores as shown in Equation \ref{eqn:BERT} as our ranking metric.  This process, carried out using the $\mathtt{distilbert-base-uncased}$ \cite{Sanh2019DistilBERTAD} model, requires approximately seven seconds to complete on each test set. 

\vspace{0.8mm}
\noindent \textbf{Lemma Similarity} (\lemma): The lemma\footnote{We use spaCy 3.4 \texttt{en\_core\_web\_md} lemmatizer} similarity method emulates the annotation process carried out by human annotators when determining coreference based on keyword comparisons between two mentions. To estimate this similarity, we compute the token overlap (Jaccard similarity; \JS) between the triggers and sentences containing the respective mentions and take a weighted average of the two similarities (like Eq \ref{eqn:BERT}) as shown in Eq \ref{eqn:lemma}\footnote{$\lambda$ is a hyper-parameter to control the weightage of the trigger and sentence similarities in Equations \ref{eqn:BERT} and \ref{eqn:lemma}, which we tune using the development set. See Appendix \ref{app:lambda}.}.
%
\begin{align}
\label{eqn:lemma}
\begin{split}
\lemma(\msubi,&\msubj) = \lambda~\JS(\tsubmi, \tsubmj)\\
                     &+ (1 -\lambda)~\JS(\Ssubmi, \Ssubmj)
\end{split}
\end{align}
%

\noindent \textbf{No Ranking} (\manual):  For our baseline approach, we employ a method that directly picks the candidate-mention pairs through random sampling and without ranking, providing a reference point for evaluating the effectiveness of the above three ranking techniques.
\vspace{-0.6mm}
\subsection{Pruning}
\vspace{-0.5mm}
To control the comparisons between candidate and target events, we restrict our selection to the top-\n~ranked candidates. To refine our analysis, we employ non-integer \n~values, allowing for the inclusion of an additional candidate with a probability equal to the decimal part of \n. We vary the values of \n~from 2 to 20 on increments of 0.5 and then investigate its relation to recall and effort in \S\ref{sec:eval}.


\vspace{-0.6mm}
\subsection{Simulation}
\vspace{-0.5mm}
To evaluate the ranking methods, we conduct annotation simulations on the events in the \ecb~and \gvc~development and test sets. These simulations follow the same annotation methodology of retrieving and ranking candidate events for each target but utilize ground-truth for clustering. By executing simulations on different ranking methods and analyzing their performance, we effectively isolate and assess each approach.
\vspace*{-0.9mm}
\section{Evaluation Methodology}
\vspace*{-0.9mm}
\label{sec:eval}
We evaluate the performance of the model-in-the-loop annotation with the ranking methods through simulation on two aspects: (1) how well it finds the coreferent links, and (2) how much effort it would take to annotate the links using the ranking method. 
\begin{figure*}[t!]
  \centering
  \hspace*{-5mm}
  \scalebox{0.85}{\input{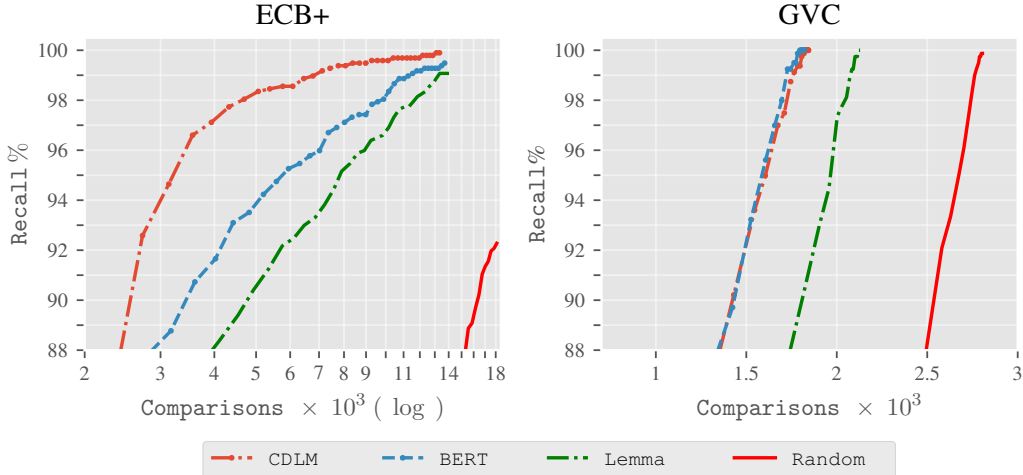}}
  \vspace*{-2.5mm}
\caption{\mucr~and \comps~achieved upon varying the \n~for each ranking method in the ECR annotation simulation. The three methods result in significantly fewer comparisons than the no-ranking \manual~baseline.}  \label{fig:recall-effort}
  \vspace*{-3mm}
\end{figure*}
\subsection{Recall-Annotation Effort Tradeoff}
\label{recall-comparisons}
\textbf{\mucr:} 
\vspace*{-0.5mm}
The recall metric evaluates the percentage of ECR links that are correctly identified by the suggestion model. It is calculated as the ratio of the number of times the true coreferent candidate is among the suggested candidates. The recall error is introduced when the coreferent candidate is erroneously removed based on the top-\n~value\footnote{Note that recall is always 100\% if no candidates are ever pruned.}.


\textbf{\comps:} A unit effort represents the comparison between a candidate and target mentions that an annotator would have to make in the annotation process. We count the sampled candidates for each target and stop counting when the coreferent candidate is found. For example, the number of comparisons for the target \mone, in Figure \ref{fig:workflow}, is 2 (\mtwo~and \mfour).  We count this number for each target event and present the sum as \comps.

\vspace*{-1.5mm}
\subsection{Analysis and Discussion}
\vspace*{-0.5mm}
\label{sec:analysis}
We present an analysis of the various ranking methods employed in our study, highlighting the performance and viability of each approach.
We employ the ranking methods on the test sets of \ecb~and \gvc. Then, estimate the \mucr~and \comps~measures for different \n~values, and collate them into the plots as shown in Figure \ref{fig:recall-effort}.

\vspace{1.mm}
\noindent\textbf{Performance Comparison:}
The performance improvement of \cdlm~ over \bert~ and \bert~ over \lemma~can be quantified by examining the graph for the ECB+ and GVC datasets. For example, when targeting a 95\% recall for the \ecb~corpus, \cdlm~provides an almost 100 percent improvement over \bert~ reducing the number of comparisons to almost half of the latter. However, both \cdlm~ and \bert~outperform \lemma~by a significant margin while being drastically better than the \manual~baseline (See Fig.~\ref{fig:recall-effort}). Interestingly, for GVC, the performance gap between \cdlm~and \bert~is quite close, both needing at least three-fourths as many comparisons as the \lemma~and crucially outperforming the \manual~baseline. \cdlm's inconsistent performance on \gvc~suggests that a corpus-fine-tuned model such as itself is more effective when applied to a dataset similar to the one it was trained on.



\vspace{1.mm}
\noindent\textbf{Efficiency and Generalizability of} \bert:

\noindent\bert~offers a compelling advantage in terms of efficiency, as it can be run on low-compute settings. Moreover, \bert~exhibits greater generalizability out-of-the-box when comparing its performance on both the ECB+ and GVC datasets. This makes it an attractive option for ECR annotation task especially when compute resources are limited or when working with diverse corpora.

\vspace*{-1.1mm}
\section{Conclusion}
\vspace*{-0.7mm}
We introduced a model-in-the-loop annotation method for annotating ECR links. We compared three ranking models through a novel evaluation methodology that answers key questions regarding the quality of the model in the annotation loop (namely, recall and effort). Overall, our analysis demonstrates the viability of the models, with \cdlm~exhibiting the best performance on the ECB+ dataset, followed by \bert~and \lemma. The choice of ranking method depends on the specific use case, dataset, and resource constraints, but all three methods offer valuable solutions for different scenarios.

\clearpage

\section*{Limitations}

It is important to note that the approaches presented in this paper have several constraints. Firstly, the methods presented are restricted to English language only, as \lemma~necessitates a lemmatizer and, \bert~and \cdlm~rely on models trained exclusively on English corpora. Secondly, the utilization of the \cdlm~model demands at least a single GPU, posing potential accessibility issues. Thirdly, ECR annotation is susceptible to errors and severe disagreements amongst annotators, which could entail multiple iterations before achieving a gold-standard quality. Lastly, the generated corpora may be biased to the model used during the annotation process, particularly for smaller values of \n.

\section*{Ethics Statement}
We use publicly-available datasets, meaning any bias or offensive content in those datasets risks being reflected in our results.  By its nature, the Gun Violence Corpus contains violent content that may be troubling for some.

\section*{Acknowledgements}
We would like to express our sincere gratitude to the anonymous reviewers whose insightful comments and constructive feedback helped to greatly improve the quality of this paper. We gratefully acknowledge the support of U.S. Defense Advanced Research Projects Agency (DARPA) FA8750-18-2-0016-AIDA – RAMFIS: Representations of vectors and Abstract Meanings For Information Synthesis. Any opinions, findings, conclusions, or recommendations expressed in this material are those of the authors and do not necessarily reflect the views of DARPA or the U.S. government. We would also like to thank ExplosionAI GmbH for partly funding this work. Finally, we extend our thanks to the BoulderNLP group and the SIGNAL Lab at Colorado State for their valuable input and collaboration throughout the development of this work.

\bibliography{anthology,custom}
\bibliographystyle{acl_natbib}

\appendix

\section{\ecb~Corpus Event Statistics}
\label{sec:corpus_details}
Table \ref{tab:ecb_gvc} contains the detailed statistics for both the \ecb \hspace{1mm} and the GVC corpora. 

	

		
    
		

\def\arraystretch{1.2}%
\begin{table}[htb]
\centering
\small
    \begin{tabular}{|c|c|c|c|c|c|c|}  \cline{2-7}
     \multicolumn{1}{c|}{~} & \multicolumn{3}{c|}{\ecb} & \multicolumn{3}{c|}{\gvc}  \\ \cline{2-7}
    \multicolumn{1}{c|}{~} & Train & Dev & Test & Train & Dev & Test \\ \hline
	T/ST   & 25 & 8 & 10 & 170 & 37 & 34 \\ \hline
	D & 594 & 196 & 206 & 358 & 78 & 74  \\ \hline
	
	M  & 3808 &  1245 & 1780 & 5313 & 977  & 1008\\ \hline

	C & 1464 & 409 & 805 & 991 & 228 &194 \\ \hline
    
    S & 1053 & 280 & 623 & 252 & 70 & 43 \\ \hline
    P & 300K & 100K & 180K & 100K & 20K & 20K \\ \hline
    P$_{+}$ & 15K & 6K & 6.5K & 24K & 3.7K & 4.1K \\ \hline
		
	\end{tabular}
  \vspace*{-2mm}
  \caption[\ecb~Corpus Statistics]{
	\ecb~ and GVC Corpus statistics for event mentions. T/ST = topics/sub-topics, D = documents, M = event mentions, C = clusters, S = singletons. P = unique mention pairs by topic. P$_{+}$ = mention pairs that are coreferent.}
\label{tab:ecb_gvc}
  \vspace*{-2mm}
\end{table}

\section{Model Details}

%
%
%

\begin{figure*}
    \centering
    \includegraphics[scale=0.75]{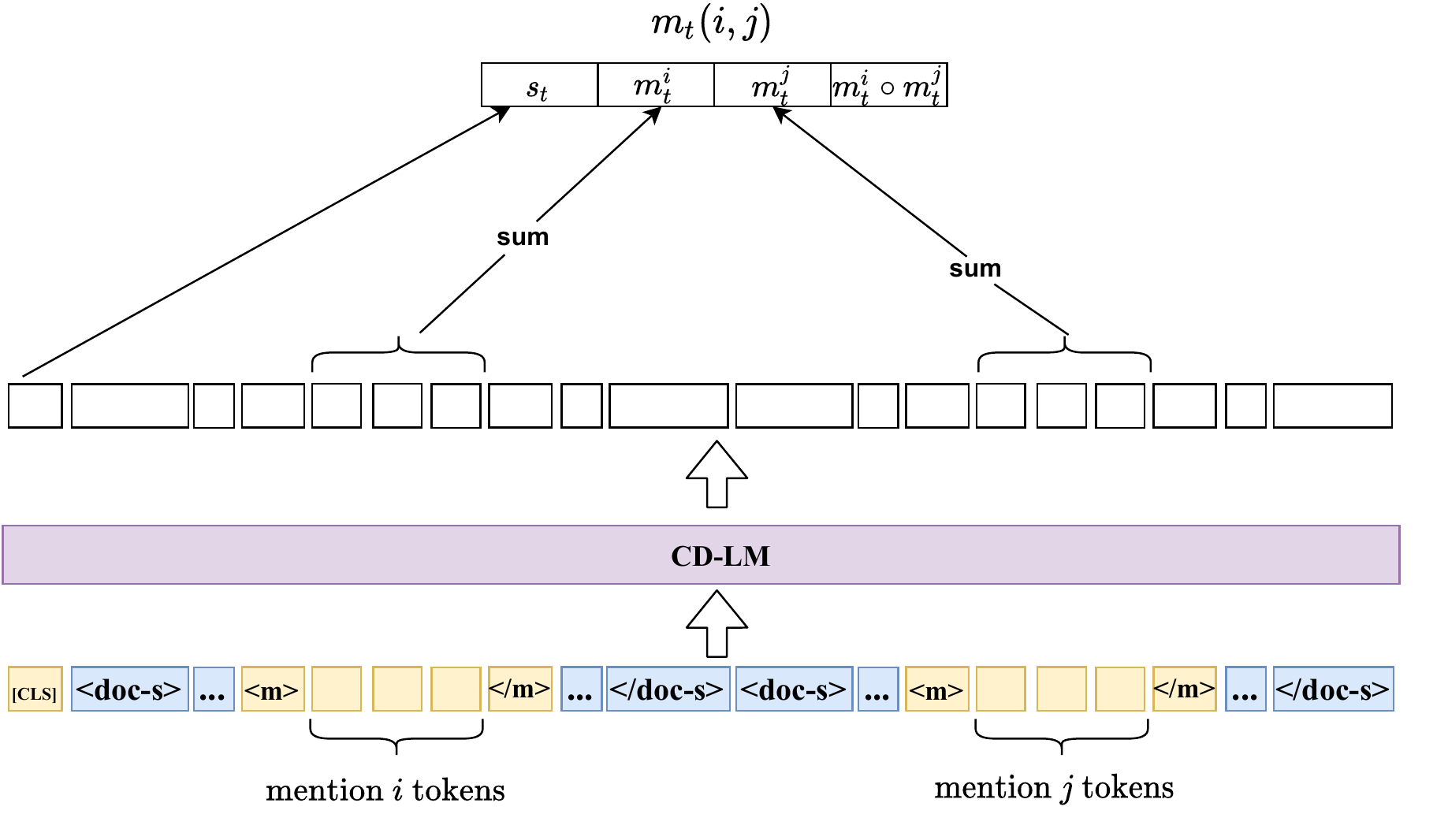}
    \caption{Illustration of Cross-encoding with CDLM from \citet{caciularu-etal-2021-cdlm-cross}.}
    \label{fig:cdlm}
\end{figure*}

\begin{figure*}[!b]
    \centering
    \includegraphics[scale=0.25]{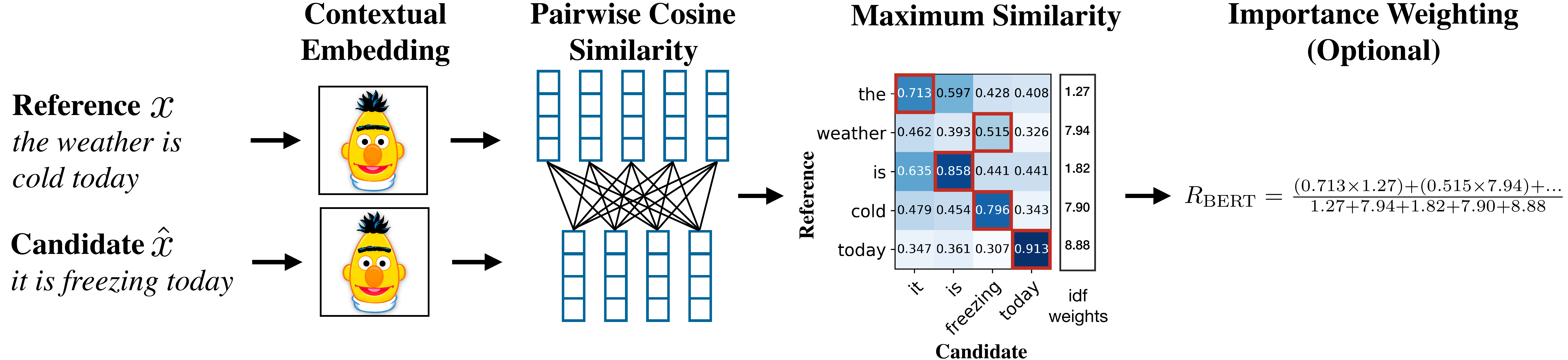}
    \caption{Illustration of the Recall Measure of BERTScore from \citet{zhang2019bertscore}.}
    \label{fig:bert-score}
\end{figure*}
\subsection{CDLM}

The CDLM model, based on the Longformer architecture, cleverly uses a combination of global and local attention for event trigger words and the rest of the document containing those events respectively. More specifically, the Longformer’s increased input capacity of 4096 tokens is utilized to encode much longer documents at finetuning that are usually seen in coreference corpora like the ECB+.  As seen in Fig.~\ref{fig:cdlm}, apart from the document-separator tokens like <doc-s> and <doc-s/> that help contextualize each document in a pair, it adds two special tokens (<m> and </m>) to the model vocabulary while pretraining to achieve a greater level of contextualization of a document pair while attending to the event triggers globally at finetuning. Apart from the event-trigger words, the fine-tuned CDLM model also applies the global attention mechanism on the [CLS] token resulting in a more refined embedding for that document pair while maintaining linearity in the transformer’s self-attention. 
\label{app:cdlm}

\subsection{BERTScore}
\label{app:bert}
BERT-Score is an easy-to-use, low-compute scoring metric that can be used to evaluate NLP tasks that require semantic-similarity matching. This task-agnostic metric uses a base language model like BERT to generate token embeddings and leverages the entire sub-word tokenized reference and candidate sentences ($x$ and $\hat x$ in Fig.~\ref{fig:bert-score}) to calculate the pairwise cosine similarity between the sentence pair. It uses a combination of a greedy-matching subroutine to maximize the similarity scores while normalizing the generated scores based on the IDF (Inverse Document Frequency) of the sub-tokens thereby resulting in more human-readable scores. The latter weighting parameter takes care of rare-word occurrences in sentence pairs that are usually more indicative of how semantically similar such pairs are. In our experiments, we use the $\mathtt{distilbert-base-uncased}$ model to get the pairwise coreference scores, consistent with our goal of deploying an annotation workflow suitable for resource-constrained settings. Such lighter and 'distilled' encoders allow us to optimize resources at inference with minimal loss in performance. 
\begin{figure*}[t]
  \centering
  \hspace*{-5mm}
  \scalebox{0.95}{\input{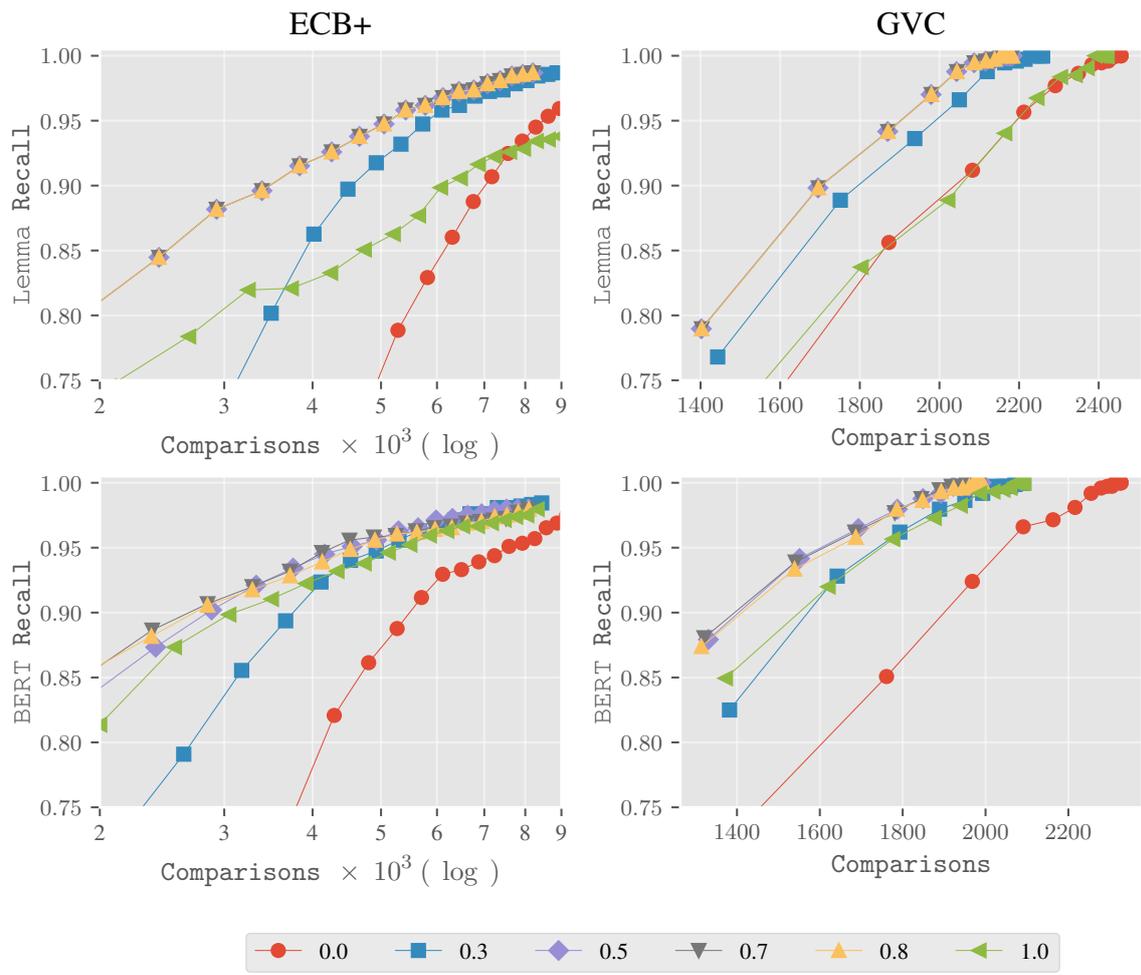}}
  \vspace*{-2.5mm}
\caption{Trigger and Sentence Similarity weight ($\lambda$) Hyper-parameter tuning on the development sets of \ecb~and \gvc. We deduce $\lambda = 0.7$~is optimal for both methods for both datasets.}  \label{fig:lambda-analysis}
  \vspace*{-3mm}
\end{figure*}
\section{$\lambda$~Hyper-parameter Tuning}
We employ the evaluation methodology detailed in \S\ref{sec:eval} to determine the optimal value of $\lambda$~(the weight for trigger similarity and sentence similarity) for both \bert~and \lemma~approaches. By conducting incremental annotation simulations on the development sets of ECB+ and GVC, we assess $\lambda$~values ranging from 0 to 1. The recall-effort curve is plotted for each $\lambda$~value, as shown in Figure \ref{fig:lambda-analysis}, allowing us to identify the one that consistently achieves the highest recall with the fewest comparisons. Remarkably, the optimal value for both methods is found to be 0.7, and this value remains consistent across the two datasets and approaches.
\label{app:lambda}

\section{Annotation Interface using Prodigy}
\label{app:prodigy}

\begin{figure*}[t]

  \centering
  \hspace*{-5mm}
  \includegraphics[scale=0.5]{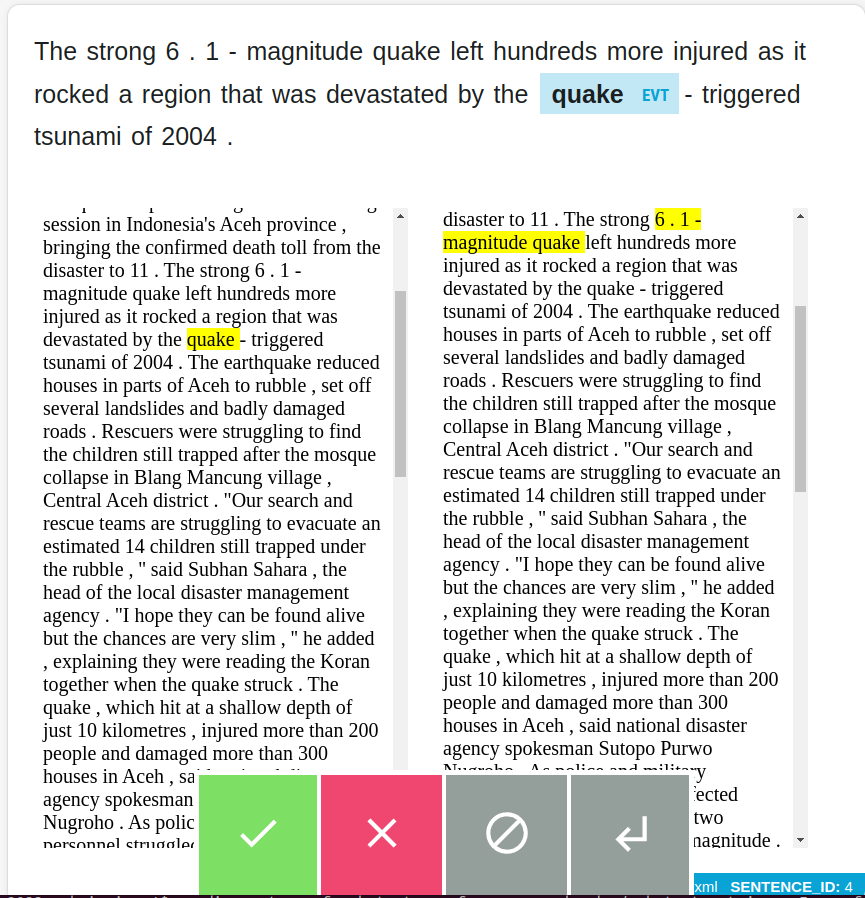}
  \vspace*{-2.5mm}
\caption{The model-in-the-loop ECR annotation using the Prodigy Annotation Tool. The target event is on the left and the Candidate cluster is on the right. }  
  \vspace*{-3mm}
  \label{fig:anno_ui}
\end{figure*}
Figure \ref{fig:anno_ui} illustrates the interface design of the annotation methodology on the popular model-in-the-loop annotation tool - Prodigy (\href{https://prodi.gy}{prodi.gy}). We use this tool for the simplicity it offers in plugging in the various ranking methods we explained. The recipe for plugging it in to the tool along with other experiment code: \href{https://github.com/ahmeshaf/model_in_coref}{$\mathtt{github.com/ahmeshaf/model\_in\_coref}$}.

\end{document}